\documentclass[lettersize,journal]{IEEEtran}
\usepackage{amsmath,amsfonts}
\usepackage{array}
\usepackage[caption=false,font=normalsize,labelfont=sf,textfont=sf]{subfig}
\usepackage{textcomp}
\usepackage{stfloats}
\usepackage{url}
\usepackage{verbatim}
\usepackage{graphicx}
\usepackage{cite}
\usepackage{orcidlink}

\usepackage{amsmath,amssymb,amsfonts}
\usepackage[ruled,vlined]{algorithm2e}
\usepackage{xcolor}
\usepackage{adjustbox}
\usepackage{multirow}
\usepackage{tikz}
\usepackage{colortbl}
\usepackage{bbding}
\usepackage{bm}
\usepackage{booktabs} 
\usepackage{pifont}
\usepackage{caption}

\newcommand{\rightsign}{\textcolor{green!60!black}{\ding{51}}} 
\newcommand{\falsesign}{\textcolor{red}{\ding{55}}} 

\definecolor{mygrey}{rgb}{ .863,  .863,  .863}

\def\BibTeX{{\rm B\kern-.05em{\sc i\kern-.025em b}\kern-.08em
    T\kern-.1667em\lower.7ex\hbox{E}\kern-.125emX}}

\begin{document}

\title{Towards Relaxed Multimodal Inputs for Gait-based \\ Parkinson's Disease Assessment}

\author{Minlin Zeng\,\orcidlink{0009-0004-2025-0901}, Zhipeng Zhou\,\orcidlink{0000-0002-1564-5800}, Yang Qiu\,\orcidlink{0000-0003-0836-782X}, Martin J. McKeown\,\orcidlink{0000-0002-4048-0817}, Zhiqi Shen\,\orcidlink{0000-0001-7626-7295}
\thanks{Minlin Zeng, Zhipeng Zhou, Yang Qiu, and Zhiqi Shen are with College of Computing and Data Science, Nanyang Technological University, 50 Nanyang Avenue, Singapore 639798 (email: minlin001, webank-zpzhou, qiuyang, zqshen{@e.ntu.edu.sg}).}
\thanks{M. J. McKeown is with the Department of Medicine (Neurology), Pacific Parkinsons Research Centre, The University of British Columbia, Vancouver, BC V6T 2B5, Canada (email: martin.mckeown@ubc.ca).}
}


\markboth{IEEE TRANSACTIONS ON CYBERNETICS}%
{Shell \MakeLowercase{\textit{et al.}}: A Sample Article Using IEEEtran.cls for IEEE Journals}


\maketitle

\begin{abstract}
Parkinson's disease assessment has garnered growing interest in recent years, particularly with the advent of sensor data and machine learning techniques. Among these, multimodal approaches have demonstrated strong performance by effectively integrating complementary information from various data sources. However, two major limitations hinder their practical application: (1) the need to synchronize all modalities during training, and (2) the dependence on all modalities during inference. To address these issues, we propose the first Parkinson’s assessment system that formulates multimodal learning as a multi-objective optimization (MOO) problem. This not only allows for more flexible modality requirements during both training and inference, but also handles modality collapse issue during multimodal information fusion. In addition, to mitigate the imbalance within individual modalities, we introduce a margin-based class rebalancing strategy to enhance category learning. We conduct extensive experiments on three public datasets under both synchronous and asynchronous settings. The results show that our framework—Towards Relaxed InPuts (\texttt{TRIP})—achieves state-of-the-art performance, outperforming the best baselines by 16.48, 6.89, and 11.55 percentage points in the asynchronous setting, and by 4.86 and 2.30 percentage points in the synchronous setting, highlighting its effectiveness and adaptability.  
\end{abstract}


\section{Introduction}
\IEEEPARstart{P}{arkinson's}  disease (PD) affects approximately 1–2\% of the global population over the age of 65 and is second only to Alzheimer’s disease among neurodegenerative disorders~\cite{brakedal2022nationwide}. PD symptoms predominantly include tremor, speech impairment, and gait disturbances, arising from motor neuron degeneration. Symptom severity increases with disease progression; therefore, timely intervention is crucial to prevent deterioration and preserve quality of life. At present, PD assessment is mainly conducted by physicians using well-established clinical scales such as the Movement Disorder Society–Unified Parkinson’s Disease Rating Scale (MDS-UPDRS) and the Hoehn and Yahr (H\&Y) scale~\cite{bloem2021parkinson}. However, in the absence of a definitive biomarker, the questionnaire-based assessment is time- and resource-intensive, subjective, and insufficiently sensitive to minor fluctuations~\cite{zhao2025artificial}. Thus, there is a substantial need to streamline PD detection and progression assessment. With the growing availability of sensing technologies, researchers can now collect high-precision motion data from PD patients, providing rich, objective information that can complement traditional in-clinic evaluation.

\begin{table}
\setlength\tabcolsep{2pt}
\footnotesize
\centering
\caption{Comparison of requirements of mainstream multimodal-based PD assessment approaches. `Single' represents single modality-based approaches, while `Early/Late Fusion', `Cross Att.', and `Shared Latent' represent the corresponding multimodal solutions.}
\label{table:comp}
\begin{adjustbox}{max width=\linewidth,center}
\begin{tabular}{lllll|l}
\toprule
          &     Single           &   Early/Late Fusion       &    Cross Att.          &   Shared Latent   &  \texttt{TRIP}  \\    \midrule
Multimodal   &     \falsesign    &   \rightsign  &    \rightsign  &  \rightsign      &    \rightsign  \\
Asynchronous &     -  &   \falsesign    &    \falsesign    &   \falsesign      &  \rightsign  \\ 
Optional Modality &     \falsesign    &   \falsesign  &    \falsesign  &  \rightsign     &    \rightsign  \\
\bottomrule
\end{tabular}
\end{adjustbox}
\end{table}

Many types of sensor data have been explored for the assessment of PD, including speech~\cite{ benba2015voiceprints}, 
movement~\cite{gao2025regular}, 
neuroimaging~\cite{wang2024diagnostic}, 
and others. Among these modalities, gait-based movement data offer several advantages for diagnosis and severity monitoring~\cite{rao2025survey}. First, gait disturbance is a prominent motor symptom even in the early stages of PD~\cite{el2020deep}. Second, quantitative gait analysis—explicitly endorsed in clinical guidelines (e.g., Timed Up and Go)—provides objective and reproducible metrics. Moreover, portable devices such as pressure sensors~\cite{williamson2021detecting}, depth sensors~\cite{zhao2021multimodal},
cameras~\cite{gupta2023new},
their combinations~\cite{creagh2020smartphone},
et al \cite{pan2024sat} can capture multi-perspective high-precision gait data, enabling a cost-effective yet multidimensional view of the pathophysiology of PD. 

On the other hand, artificial intelligence (AI) algorithms, in particular, have been extensively applied in the existing literature to recognize disease-specific patterns stored in these gait data to aid in the evaluation of PD \cite{rao2025survey}. The current literature can be categorized into single- and multi-modality ones, where single-modality approaches rely on a single data source, 
while multimodal approaches integrate multiple data streams (e.g., combining camera and inertial sensors\cite{zhao2021multimodal}) for a more comprehensive assessment. 
Although single-modality methods have shown competitive performance and currently dominate the field, an increasing body of work has demonstrated the advantages of multi-modality data in PD analytics~\cite{zhao2025artificial}, contributing to the growing popularity of multimodal solutions. Consequently, in this work we focus on multimodal gait data for PD assessment.

However, existing multimodal solutions rely on a strict data collection process, requiring synchronized inputs (i.e., strictly time-aligned signals) from full modalities during both training and inference (see Table~\ref{table:comp}). This requirement presents several key challenges, including technical and personnel-related barriers. Technically, capturing synchronized multimodal gait data (e.g., vertical ground reaction force (vGRF) and skeleton data) often demands specialized equipment and complex, non-standardized experimental setups. Personnel-related issues include data labeling and calibration for synchronizing input data\cite{zhang2022multimodal}. Moreover, collecting all modalities during inference is often impractical due to privacy concerns and device limitations\cite{bavli2025ethical}. Therefore, an approach that handles missing modality and asynchronous streams can greatly alleviate these constraints.


\begin{table}[t]
  \centering
  \caption{Ablation accuracies (\%) under synchronous input conditions. The two datasets, WearGait and FOG, consist of two classes with three modalities and three classes with two modalities, respectively. The modality and its results indicating modality collapse are highlighted in grey.}
  \label{tab:modality_collapse}
  \begin{adjustbox}{max width=\linewidth, center}
  \begin{tabular}{llcccc}
    \toprule
    Dataset & Modality & Model 1 & Model 2 & Model 3 \\
    
    \midrule
    \multirow{6}{*}{\begin{tabular}[c]{@{}c@{}}\textit{WearGait}\\ \textit{Dataset}\end{tabular}} & \multicolumn{1}{c}{} & \multicolumn{1}{c}{Early Fusion} & \multicolumn{1}{c}{Late Fusion} & \multicolumn{1}{c}{Cross Attention} \\
    \cmidrule(lr){3-5}
    &w/o Insole          & 60.02 & 53.41 & 56.39  \\
    &w/o Walkway         & 61.72 & 56.84 & 52.01  \\
    &\cellcolor{mygrey}\textbf{w/o IMU}   & \cellcolor{mygrey}\textbf{50.72} & \cellcolor{mygrey}\textbf{52.38} & \cellcolor{mygrey}\textbf{50.49}  \\
    &Full Modality       & 82.44 & 71.36 & 77.20  \\
    
    \midrule
    \multirow{5}{*}{\begin{tabular}[c]{@{}c@{}}\textit{FOG}\\ \textit{Dataset}\end{tabular}} & \multicolumn{1}{c}{} & \multicolumn{1}{c}{Early Fusion} & \multicolumn{1}{c}{Late Fusion} & \multicolumn{1}{c}{Shared Latent} \\
    \cmidrule(lr){3-5}
    &\cellcolor{mygrey}\textbf{w/o Sensor}               & \cellcolor{mygrey}\textbf{36.42} & \cellcolor{mygrey}\textbf{36.71}  & \cellcolor{mygrey}\textbf{36.14} \\
    &w/o Skeleton             & 40.09 & 47.64  & 43.56 \\
    &Full Modality            & 61.65 & 63.71 & 61.13 \\
    \bottomrule
  \end{tabular}
  \end{adjustbox}
\end{table}


In addition, during our experiments with some multimodal fusion models we observed a recurring failure mode which causes the fused model heavily leans on one subset of modalities and if this subset of modalities are removed during inference, the model's performance significantly degraded (see Table. ~\ref{tab:modality_collapse}). This issue has serious effect on our scenario as missing modality is very common in real-clinic. In the synchronous setting, ablations make this asymmetry explicit: on WearGait, Early Fusion falls from 82.44\% (full) to 50.72\% when IMU is removed, yet remains 60.02\%/61.72\% after removing Insole/Walkway; on FOG, Early/Late/Shared drop to 36.42/36.71/36.14\% without Sensor (near 33\% chance for three classes), but stay at 40.09/47.64/43.56\% without Skeleton. To the best of our knowledge, this phenomenon has not been discussed in gait-based PD assessment. Nevertheless, a subsequent review showed it is not unique to our field and has been discussed in multimodal computer vision field as “\textit{modality collapse}”, where fusion models rely on a subset of modalities or suppress one branch during optimization~\cite{chaudhuricloser}.  

To address above mentioned challenges encountered frequently in real-clinic scenarios, we propose a modality-relaxed multimodal framework that allows asynchronous and optional modality inputs during both training and inference. As a first step, we design a new architecture (see Fig.~\ref{fig:framework_train_vs_test}) which enables interaction between modalities while preserving modality-specific features.
Next, we employ a multi-objective optimization (MOO) algorithm that balances convergence and conflict avoidance to not only facilitate the learning of shared representations across modalities 
and subjects, but also mitigating the modality collapse issue.
In addition, we introduce a class-rebalanced training scheme to mitigate class imbalance within each modality. Extensive experiments are conducted under both synchronous and asynchronous conditions. The experiment results show that our model not only achieves good accuracy under asynchronous condition, but also improves single modality accuracy compares to their single-modal baselines.
In a nutshell, our contributions can be summarized as three-fold:
\begin{itemize}
    \item To the best of our knowledge, we are the first to propose a practical PD assessment system that accommodates asynchronous modality inputs during training and allows optional modality inputs during inference.
    \item A MOO framework is developed from both architectural and optimization perspective to learn modality-shared features and help alleviate the modality collapse issue. In addition, a margin-based rebalancing strategy is proposed to promote balanced learning within each modality.
    \item Extensive experiments on three public datasets and two different input modes demonstrate that \texttt{TRIP} not only surpasses single-modality baselines but also outperforms conventional and more advanced multi-modal fusion strategies, highlighting the effectiveness and flexibility of our approach.  
\end{itemize}

\section{Related Work}
Our work primarily lies at the intersection of PD assessment and MOO. In the following, we introduce key classical and recent developments in both domains and then explicitly highlight the connections and differences between our approach and existing research.

\subsection{Gait-Based Single-Modality Solutions}

Advancements in sensor devices and AI have enabled the shift in PD assessment from in-clinic evaluations with specialist supervision to AI-enhanced, automated, or remote monitoring approaches \cite{ gholami2023automatic, zhou2024sample, feng2024enhancing}. 
A common direction is to analyze gait data during locomotion, which includes kinematic data (e.g., RGB-D video \cite{zhao2021multimodal},
IMU \cite{fernandes2018artificial}) and kinetic data (e.g., vGRF \cite{zhao2018hybrid}). Early ML-based approaches use hand-crafted features (e.g., stride, speed, and cadence) with algorithms such as SVM \cite{vidya2021gait}, k-NN \cite{zhao2022severity}, or ensembles \cite{balaji2020supervised}, but require extensive manual feature engineering. To reduce this reliance, DL models are introduced to learn representations directly from raw data \cite{abumalloh2024parkinson}, ranging from MLPs \cite{zeng2019classification} to CNNs \cite{kaur2022vision}, LSTMs \cite{vidya2022parkinson}, and GNNs \cite{zhong2022robust}, each suited for spatial or temporal signals. More recent methods adopt transformers\cite{xia2019dual} and hybrid models\cite{zhao2018hybrid} to capture complex spatiotemporal patterns. Overall, DL has been shown to consistently outperform ML in many PD tasks \cite{skaramagkas2023multi}.

\subsection{Gait-Based Multi-Modality Solutions}

In addition to single-modality studies, multimodal approaches have demonstrated superior capacity in capturing holistic disease patterns by integrating diverse data sources \cite{zhao2025artificial}. However, effectively utilizing cross-modal information remains challenging. Early works either relied on ML with hand-crafted features \cite{radu2018multimodal} or combined modality-specific architectures (e.g., 3D CNNs and LSTMs \cite{kumar2018multimodal, castro2020multimodal}), to extract spatial and temporal features separately. While these methods capture partial spatiotemporal or multimodal characteristics, they struggle with cross-modal correlation. More recent efforts can be categorized into three strategies (see Table \ref{table:comp}): early/late fusion, cross-attention, and shared latent representation. For instance, \cite{zhao2021multimodal} proposed a late fusion method that uses separate spatial encoders followed by Correlative Memory Neural Networks to learn joint temporal embeddings. The work of \cite{hu2021graph} introduced a graph-based shared latent fusion by modeling each modality as graph vertices to learn inter-modality relationships. Another work \cite{cui2023multi} adopted cross-attention modules, applying bi-directional co-attention across silhouette and skeleton streams to extract complementary information.

Despite their effectiveness, these models face practical limitations. Most require either strict spatial-temporal alignment or complete modality availability during inference—constraints that are difficult to meet due to the complexity and cost of collecting fully synchronized multimodal gait data.

\subsection{Multi-Objective Optimization}
Multi-objective optimization (MOO) has been proposed to address machine learning problems involving multiple objectives. It can generally be categorized into two types: 1. Pareto Front Learning (PFL) and 2. Balanced Trade-off Exploration. PFL aims to approximate the entire Pareto front so that the desired trade-off can be achieved once user preferences are specified. In this direction, several methods have been introduced in recent years. PHN~\cite{navon2020learning} uses a hypernetwork to generate Pareto-optimal models conditioned on preference vectors. COSMOS~\cite{ruchte2021scalable} reduces the parameter overhead by conditioning in the feature space rather than using a hypernetwork. PaMaL~\cite{dimitriadis2023pareto} learns the Pareto front in the manifold space by optimizing task-specific endpoints. To improve parameter efficiency, follow-up works have proposed low-rank approximations~\cite{dimitriadis2024pareto,chen2024efficient} and mixture-of-experts (MoE) architectures~\cite{tang2024towards} to reduce endpoint overhead.

On the other hand, balanced trade-off exploration is commonly used in multi-task learning (MTL), which seeks to achieve balanced progress across tasks. Various gradient-based MOO methods have been proposed in this context. MGDA~\cite{sener2018multi} applies the Frank–Wolfe algorithm to find the gradient combination with the smallest norm. PCGrad~\cite{yu2020gradient} mitigates gradient conflicts by projecting gradients onto orthogonal subspaces. CAGrad~\cite{liu2021conflict} balances convergence and conflict avoidance through compromise objectives. Nash-MTL~\cite{navon2022multi} adopts a game-theoretic approach, allowing tasks to negotiate parameter updates. Building on this, FairGrad~\cite{ban2024fair} introduces a finer-grained constraint to ensure equitable learning progress among tasks.

\noindent \textbf{Connection and Difference}:
Similar to prior work, our approach also leverages multimodal information to enhance PD assessment. However, as summarized in Table~\ref{table:comp}, a key distinction is that our proposed MOO framework eliminates the requirement for synchronous multimodal inputs during both training and inference, thereby improving practicality and aligning better with real-world deployment scenarios. To the best of our knowledge, the most related work is that of Heidarivincheh et al.~\cite{heidarivincheh2021multimodal}, which employs a variational autoencoder (VAE) to encode all modalities into a shared latent space, allowing for optional modality input at inference. Nevertheless, their method does not support asynchronous multimodal input during training, a limitation that our work explicitly addresses.
\begin{figure*}[ht]
    \centering
    \includegraphics[width=\textwidth, trim=0.7cm 6cm 0.6cm 6.3cm, clip]{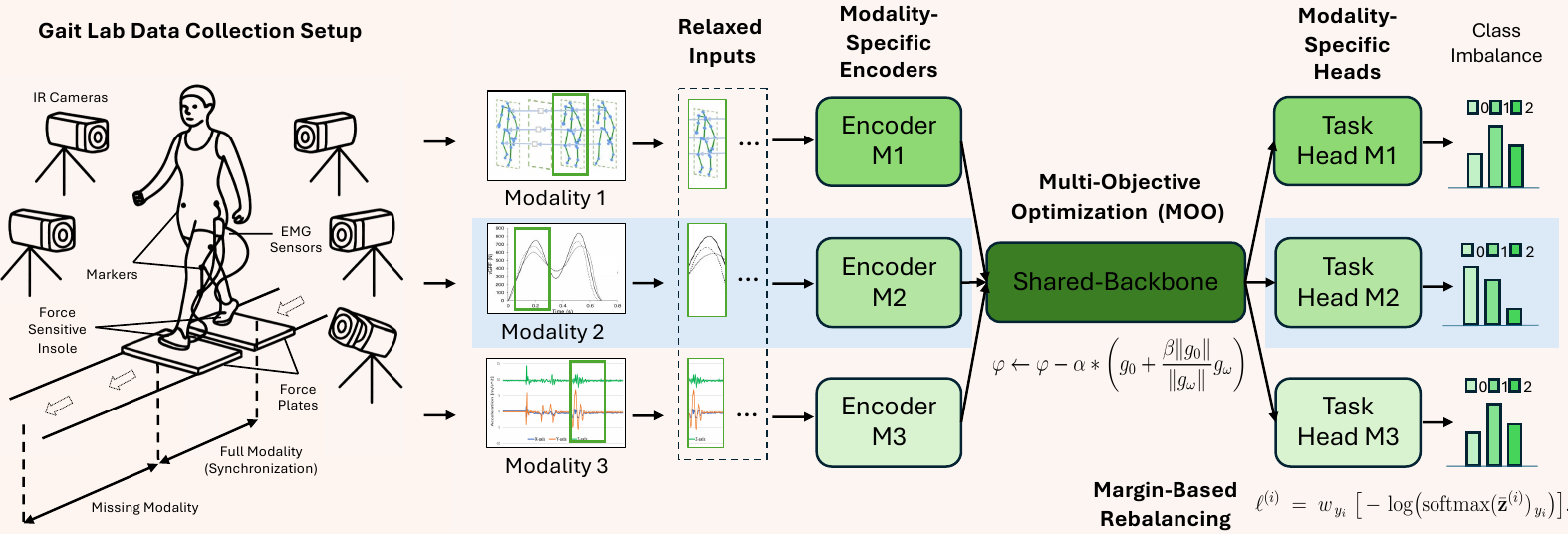}
    \caption{\texttt{TRIP} overview (relaxed-input scenario).  Gait lab setup: conventional labs record multiple synchronized modalities (e.g., 3D motion capture and wearables), but trials often have missing streams or dropouts. Our pipeline supports asynchronous inputs and missing modalities through: (1) \textbf{Relaxed Inputs}: trains and infers with time-misaligned clips and enabled missing modalities (blue bands); each stream is encoded independently, then processed by a shared backbone and modality heads—matching real deployment. (2) \textbf{Shared-Only MOO}: modality losses update their own private parameters, while only the shared parameters $\varphi$ receive mixed, conflict-averse gradients using our proposed MOO objective. (3) \textbf{Imbalance-Aware Heads}: our margin-based rebalancing strategy at each head reweights classes with cosine-normalized logits to handle the inter/intra-class imbalance typical of multimodal gait PD data.}
    \label{fig:framework_train_vs_test}
\end{figure*}

\section{Principal Design}
In this section, we present a detailed overview of our proposed framework (see Fig. \ref{fig:framework_train_vs_test}), architectural design, MOO learning paradigm, and class rebalancing strategy.

\subsection{Problem Setup}

We consider multimodal gait-based PD assessment with \(m\) input modalities per subject. Let
\[
\mathcal{D}=\bigl\{(x^{i}_{1},x^{i}_{2},\dots,x^{i}_{m},\, y^{i})\bigr\}_{i=1}^{N},
\]
where \(x^{i}_{r}\in\mathbb{R}^{T_r\times D_r}\) denotes the sequence window from modality \(r\in\{1,\dots,m\}\) and \(y^{i}\in\{1,\dots,K\}\) is the clinical label. At training and inference, a subset of modalities \(\mathcal{S}_i\subseteq\{1,\dots,m\}\) may be available due to asynchrony or dropout. A model \(f_{\Theta}\) consumes any available subset and predicts
\[
f_{\Theta}\!\bigl(\{x^{i}_{r}\}_{r\in\mathcal{S}_i}\bigr)\;\rightarrow\;\hat y^{i}.
\]


\subsection{Overall Framework}
\label{overall_framework}
For each modality \(r\), a modality-specific encoder \(e_r(\cdot;\,\omega_r)\) maps the raw sequence to a fixed-width feature sequence
\[
u_{r}^{\,i}=e_r\!\bigl(x_{r}^{\,i};\,\omega_r\bigr).
\]
Each \(u_{r}^{\,i}\) is then processed by a \emph{shared} encoder \(g(\cdot;\,\varphi)\) to produce shared representations
\[
s_{r}^{\,i}=g\!\bigl(u_{r}^{\,i};\,\varphi\bigr).
\]
A modality-specific head \(h_r(\cdot;\,\theta_r)\) produces logits
\[
z_{r}^{\,i}=h_r\!\bigl(s_{r}^{\,i};\,\theta_r\bigr),
\]
and a per-modality supervised loss \(\ell_r^{\,i}=\ell\!\bigl(z_{r}^{\,i},y^{i}\bigr)\) is computed for every \(r\in\mathcal{S}_i\).
Private parameters \(\{\omega_r,\theta_r\}\) are updated using their respective gradients from \(\ell_r\).
Gradients on the shared encoder \(\varphi\) induced by \(\{\ell_r\}_{r\in\mathcal{S}_i}\) are combined by a MOO step to yield a single update for \(\varphi\).
At inference, any subset of modalities can be provided.


\subsection{Architecture Design}
We adopt a modular \emph{encoders → shared backbone → task heads} design. It supports relaxed inputs at test time and cleanly separates private vs.\ shared parameters for multi–objective training. (1) \textit{Encoders}. Each modality uses a modality-specific encoder to convert raw signals into embeddings with a common feature width to interface with the same backbone.
(2) \textit{Shared Backbone}. A single backbone is reused across all streams, which transforms each encoder’s embedding into a compact representation for classification. The same weights are shared across modalities to encourage cross–modal regularization. (3) \textit{Task Heads.} Backbone features are flattened and routed to classification heads. Synchronous inputs use one shared head referenced by all streams; asynchronous inputs use one head per modality to permit training and inference with any subset of available data. Based on training conditions, private parameters are the encoders and per–modality heads in asynchronous condition; shared parameters are the backbone and the shared head in synchronous condition. 

However, while this design allows flexible inputs at inference and clearly separates modality-specific (private) parameters, conflicts can arise during training on the shared parameters (see Fig. \ref{fig:CAGrad_gradient}, left). Therefore, a strategy that resolves these potential conflicts is crucial for stable training and smooth gradient updates.


\subsection{MOO-Based Multimodal Learning}
To further enable the asynchronous and optional modality input, we develop a MOO learning paradigm to facilitate the shared feature extraction on shared backbone ($f_{\varphi}$) across modalities. 

Assume the losses from all modality are $\{\mathcal{L}_i\}_{i=1}^M$ and their derived gradients on $\varphi$ are $\{g_i\}_{i=1}^M$, where $M$ is the number of modality (set as 2 or 3 in this paper). Since our objective is to promote all modalities' learning rather than optimizing the average ones, a worst case optimization is adopted as follow:
\begin{align} \label{eqn:obj}
    \underset{d}{\rm min}\ \underset{i}{\rm max} \frac{1}{\alpha}(\mathcal{L}_i(\varphi - \alpha * d) - \mathcal{L}_i(\varphi))   
\end{align}
where $d$ is the update vector for $\varphi$, i.e., $\varphi \gets \varphi - \alpha * d$, and $\alpha$ is the learning rate. This formulation first seeks the least progress one, and then optimize $d$ to improve it. Note that Eqn.~\ref{eqn:obj} can be further transformed as follow by applying first-order Taylor approximation:
\begin{align} \label{eqn:obj_2}
    \underset{d}{\rm min}\ \underset{i}{\rm max} -g_i^{\top}d    
\end{align}
where $g_i$ is $\nabla \mathcal{L}_i(\varphi)$. Note that $\underset{i}{\rm max} -g_i^{\top}d = \underset{\omega \in [M]}{\rm max} -(\sum_i {\omega_ig_i})^{\top}d$, where $[M]$ is a simplex. Then we impose a norm constraint on $d$ in Eqn.~\ref{eqn:obj_2}, i.e., $\|d\| - \|g_0\| \le \|d - g_0\| \le  \beta\|g_0\|$. $g_0$ is the average gradient of all modalities, which serves as a na\"ive optimization. $\beta \in [0,1)$ is a hyper-parameter measures the allowed deviation from $g_0$. This constraint ensures the convergence of MOO algorithm. Therefore, we formulate the dual objective according to Lagrangian equation and Slater's condition as follow:
\begin{align} \label{eqn:obj_3}
    \underset{\omega \in [M]}{\rm min}\ \underset{d}{\rm max} \ g_{\omega}^{\top}d - \lambda (\|d - g_0\|^2 - \beta\|g_0\|^2),\ \ \ \  \lambda > 0
\end{align}
where $g_{\omega} = \sum_i {\omega_ig_i}$. By fixing $\lambda$ and $\omega$, we can obtain the optimal $d = g_0 + g_{\omega} / \lambda$. Inserting the optimal $d$ into Eqn.~\ref{eqn:obj_3}, we have:
\begin{align} \label{eqn:obj_4}
    \underset{\omega \in [M],\ \lambda}{\rm min}\ g_{\omega}^{\top}g_0 + \frac{1}{2\lambda} \|g_{\omega}\|^2 + \frac{\lambda}{2}\beta^2\|g_0\|^2
\end{align}
By applying the mean value inequality, we have the final objective as follow (when $\lambda = \frac{\|g_{\omega}\|}{\beta \|g_0\|}$):
\begin{align} \label{eqn:obj_5}
    \underset{\omega \in [M]}{\rm min}\ g_{\omega}^{\top}g_0 + \beta \|g_0\|\|g_{\omega}\|
\end{align}
Once the optimal $\omega^*$ are derived according to Eqn.~\ref{eqn:obj_5}, we have $d = g_0 + \frac{\beta\|g_0\|}{\|g_{\omega}\|}g_{\omega}$, and $\varphi$ is updated as follow:
\begin{align} \label{eqn:updated}
    \varphi \gets \varphi - \alpha * \left (g_0 + \frac{\beta\|g_0\|}{\|g_{\omega}\|}g_{\omega} \right ) 
\end{align}

\begin{figure}[t]
    \centering
    \includegraphics[height=5cm,width=\linewidth,keepaspectratio]{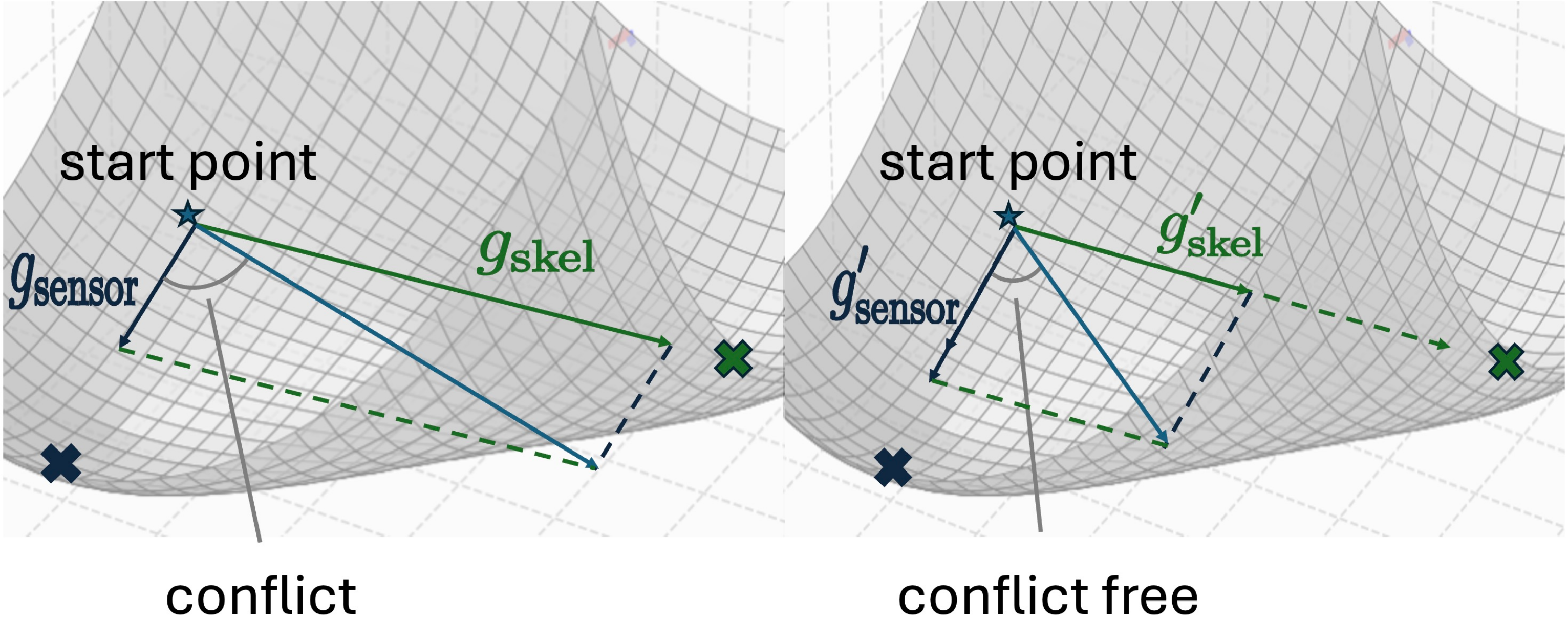}
    \caption{Illustration of the MOO learning paradigm (two-modalities scenarios). Left: gradient conflict between the update direction of the sensor modality ($g_{\text{sensor}}$) and the overall average gradient direction. Right: conflict resolved after applying MOO optimization, aligning the adjusted modality gradients ($g'_{\text{sensor}}$, $g_{\text{skel}}$) with the shared descent direction.
    }
    \label{fig:CAGrad_gradient}
\end{figure}

This process is illustrated in Fig.~\ref{fig:CAGrad_gradient}. The employed MOO algorithm adjusts the weighting of modality-specific gradients to promote conflict-averse and balanced progress across individual tasks.

\noindent \textbf{Remarks:} Only shared backbone's parameters $\varphi$ are updated via Eqn.~\ref{eqn:updated}, other modules are updated by their corresponding modality-specific gradients (see Algorithm~\ref{alg:algorithm1}). 

\subsection{Margin-Based Rebalancing Strategy}
\begin{figure}[t]
    \centering
    \includegraphics[height=4cm,width=\linewidth]{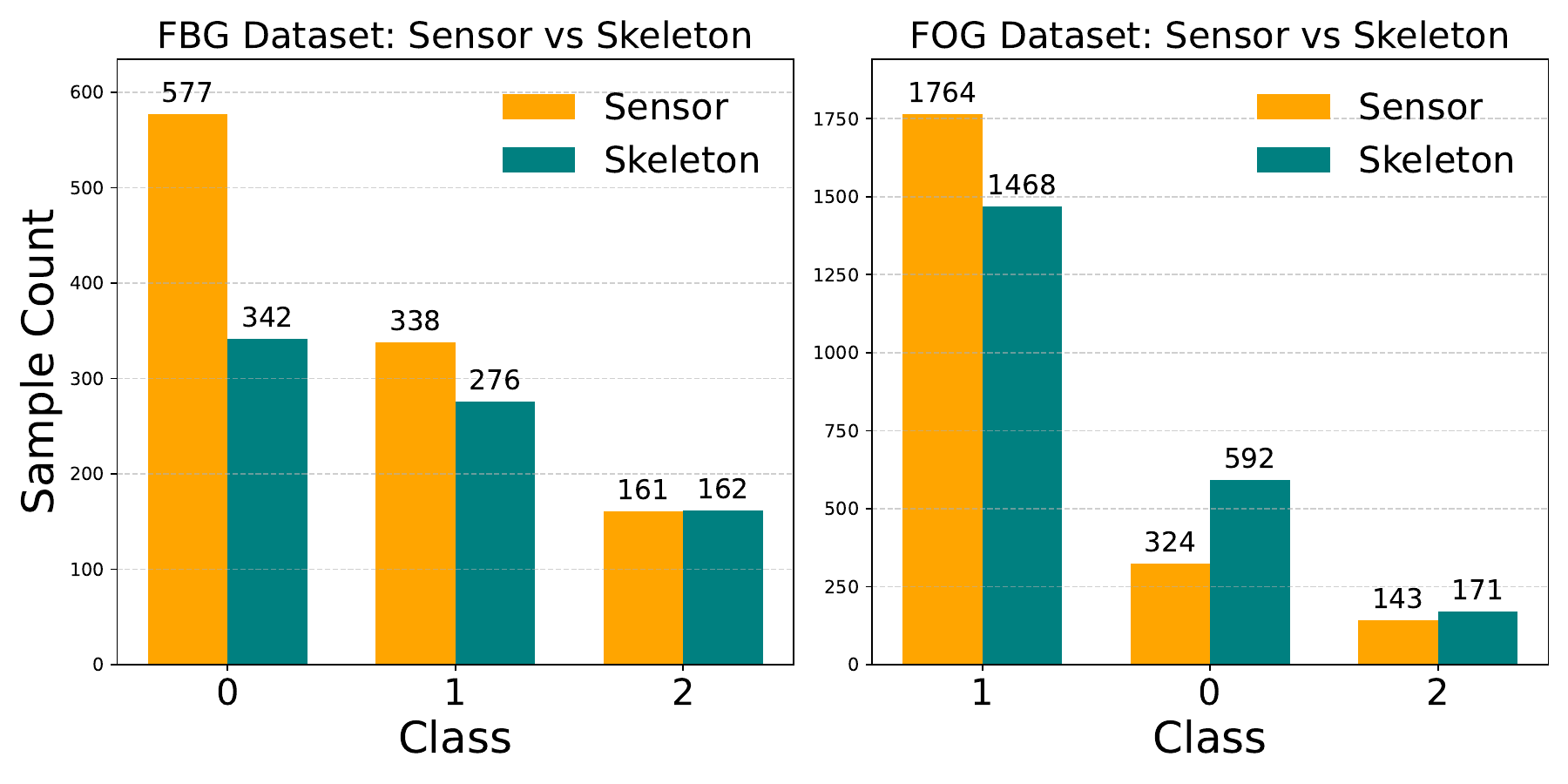}
    \caption{Class distribution of multimodal PD datasets (FBG and FOG). Both datasets exhibit long-tailed distributions and noticeable intra-class imbalance between sensor and skeleton modalities.}
    \label{fig:class_imbalance}
\end{figure}

Another issue lies in the prevalent class imbalance in PD multimodal datasets. As depicted in Fig. \ref{fig:class_imbalance}, both FOG and FBG exhibit a long-tailed distribution across categories. To mitigate the adverse effects of long‐tailed class distribution inherent in PD multimodal gait classification tasks, we integrate class‐adaptive margins and stochastic smoothing into the standard cross‐entropy framework for both modalities.

Concretely, let K denote the total number of classes and let \(N_j\) be the number of training samples for class \(j\).  We first construct a \emph{per‐class margin vector} \(\mathbf{m} = (m_1,\dots,m_C)^\top\) whose entries are proportional to \(\log(N_{\max}) - \log(N_j)\), where \(N_{\max} = \max_{k}N_k\).  That is,
\begin{align}
m_j \;=\; \log\bigl(N_{\max}\bigr) \;-\; \log\bigl(N_j\bigr),
\qquad j = 1,\ldots,K.
\end{align}
During the forward pass, suppose a given batch of size \(B\) produces cosine‐normalized logits \(\mathbf{z}^{(i)} = \bigl(z^{(i)}_{1},\dots,z^{(i)}_{K}\bigr)\) for (i=1,\ldots,B).  We then inject a small, class‐scaled random perturbation \(\delta^{(i)}_{j}\sim \mathcal{N}(0,\sigma^2)\) (clamped to \([-1,1]\)) into each logit according to
\begin{align}
    \tilde{z}^{(i)}_{j} \;=\; z^{(i)}_{j}
\;-\;\eta \,\bigl|\delta^{(i)}_{j}\bigr|\;\frac{m_j}{\max_{k}m_k},
\end{align}
where \(\eta>0\) is a noise‐magnitude hyperparameter that ensures more perturbation for under‐represented classes.  Next, we apply an \emph{additive margin} \(m\) only to the target‐class entry: if \(y_{i}\) is the ground‐truth label for sample \(i\), then
\begin{align}
\hat{z}^{(i)}_{j} \;=\;
\begin{cases}
\displaystyle \tilde{z}^{(i)}_{\,y_i} \;-\; m, 
& j \;=\; y_{i}, \\[8pt]
\displaystyle \tilde{z}^{(i)}_{j}, 
& j \;\neq\; y_{i}.
\end{cases}
\end{align}

In addition, we incorporate \emph{logarithmically scaled class weights} to compensate for imbalance.  Define
\begin{align}
w_j \;=\; \frac{\log\bigl(N_{\max}/N_j + \varepsilon\bigr)}{\mathrm{div}},
\qquad j = 1,\ldots,K,
\end{align}
where \(\varepsilon>0\) prevents \(\log(0)\) and \(\mathrm{div}\) is a tunable divisor that controls the overall weight magnitude.  We then normalize \(\mathbf{w} = (w_1,\dots,w_C)\) so that \(\sum_{j=1}^C w_j = C\).  The final loss for sample \(i\) is
\begin{align} \label{eqn:gcl}
\ell_{\mathrm{}}^{(i)}
\;=\; w_{\,y_i}\; \bigl[\,-\,\log\bigl(\mathrm{softmax}(\bar{\mathbf{z}}^{(i)})_{\,y_i}\bigr)\bigr].
\end{align}
Averaging \(\ell_{\mathrm{}}^{(i)}\) over the batch yields the overall loss.  This combination of \emph{(i)} class‐adaptive margin subtraction, \emph{(ii)} logarithmic re‐weighting, and \emph{(iii)} noise smoothing directs the model’s capacity toward underrepresented gait classes while maintaining stability on majority classes. The overall training scheme is summarized in Algorithm~\ref{alg:algorithm1}.

\subsubsection{Integrated Loss Function}
\begin{algorithm}[t]
	\caption{Training Paradigm of \texttt{TRIP}}
	\label{alg:algorithm1}
	\KwIn{Training Dataset $\mathcal{D}=\bigl\{(x^{i}_{1},\dots,x^{i}_{m},\, y^{i})\bigr\}_{i=1}^{N}$}  
	\KwOut{Model trained with \texttt{TRIP}} 
	\BlankLine
	\textbf{Stage 1}: \\
	Initialize \{$\omega_{r}$, $\varphi$, $\theta_{r}$\} randomly, with \(r\in\mathcal{S}_i\) and \(\mathcal{S}_i\subseteq\{1,\dots,m\}\)\\
	\While{\textnormal{not converged}}{
		\ForEach{batch $\mathcal{B}_i$ in $\mathcal{D}$}{
		Compute modality-specific loss $\ell_{r}$ via Eqn.~\ref{eqn:gcl}. \\
		Derived gradients on \{$\omega_{r}$, $\varphi$, $\theta_{r}$\} with respect to  $\ell_{r}$.\\
		Update $\varphi$ with $g_{\varphi}^{r}$ via Eqn.~\ref{eqn:updated}. \\
        $\omega_{r} \gets \omega_{r} - \alpha * g_{\omega_{r}}$,\\
        $\theta_{r} \gets \theta_{r} - \alpha * g_{\theta_{r}}$.
		}
    }
\end{algorithm}

\section{Implementation}
We evaluate single-modality and multimodal fusion baselines on three gait datasets for PD, using subject-wise stratified cross-validation (CV). All models share the same classification setup for fair comparison. 

\subsection{Setup}
\paragraph{Sampling Strategy}
To mitigate imbalances, we use two complementary strategies. (1) \textit{Modality-Balanced Training} (asynchronous only): in asynchronous training, the sample size of each modality varies, so we perform modality-level oversampling. However, this does not guarantee a class-balanced training set within each modality. To address this, we propose a margin-based class rebalancing strategy, which helps ensure class-balanced training. This stabilizes gradient contributions and enhances joint representation learning alongside the MOO objective. (2) \textit{Class- and Modality-Balanced Evaluation} (synchronous and asynchronous): for fair validation, we oversample the evaluation set to equalize class counts and segment counts within and across modalities.
\paragraph{Training and Hyperparameters}
Hyperparameter tuning is performed per dataset and model configuration, optimizing for classification accuracy. All models are trained using the Adam optimizer for a maximum of 100 epochs and the learning rate is fixed throughout. Experiments are conducted using an NVIDIA Tesla V100 GPU.
\subsection{Training Strategy} We use two different training strategies in
this work as detailed below. (1) \textbf{Synchronous}: this setting is commonly set as default in previous models which requires inputs to be time-aligned and from the same subject during training.
(2) \textbf{Asynchronous}: this setting is designed to simulate real-world deployments where modality dropouts and time misalignment are common at inference. During training, inputs from different modalities may come from different subjects and from different time segments. At inference, when a modality is missing, the model processes the available modality. This increases usable data and diversity through relaxing strict alignment. It also reflects clinical reality (missing/noisy streams) and encourages subject-invariant representations by exposing the shared encoder to cross-subject variability during training, while evaluation remains subject-wise to avoid leakage. 

Crucially, the asynchronous setting is ill-posed for the baseline models as mixing subjects introduces cross-label contamination. We report their asynchronous results (see Table~\ref{tab:main_results}) only for completeness—to highlight this limitation and to compare against our design, which is expressly built for asynchronous inputs.

\subsection{Validation Strategy} 

\paragraph{Cross Validation}
We use subject-wise stratified $k$-fold CV. Subjects are grouped by class; each fold’s evaluation set contains one subject per class (so $k$ is bounded by the smallest class). For multimodal experiments, only subjects with complete data in both modalities are eligible for the evaluation set; single-modality baselines reuse the same evaluation subjects. Remaining subjects form the training set. This prevents leakage, preserves train/eval independence, and yields balanced estimates.

\paragraph{Baselines}We evaluate on both single modality and multimodal fusion baselines. All models inherit similar design of encoders, backbone, and training hyper-parameters as our pipeline to isolates fusion effects from all other factors under both synchronous and asynchronous settings. \textbf{Single-Modality Baselines}: They use the exact architecture from our main pipeline, but only one encoder is active at a time. These serve as upper bounds and isolate modality–specific signal. \textbf{Conventional Methods}: They serve as naive fusion baselines for our approach.
(1) \textit{Early/Late Fusion}: Early Fusion concatenates low-level features from encoders and feeds the joint sequence to the shared backbone; Late Fusion processes each modality independently through the backbone, then concatenates the resulting high-level vectors for prediction. (2) \textit{Shared-Latent Fusion}: Each modality is linearly projected into a common latent width and fused by element-wise addition before the backbone. (3) \textit{Cross-Attention Fusion}: A lightweight, symmetric cross-attention block lets each modality attend to the other; the fused sequence is then passed to the backbone.

\textbf{SOTA Variants}: These are implemented to benchmark our method against newer designs. 
(1) \textit{FOCAL}~\cite{liu2023focal}: This method factorizes each modality into shared and private latents, enforces orthogonality between them, and applies a contrastive objective to align shared parts across modalities, preserving complementary cues while reducing single-stream shortcuts.
(2) \textit{TACA}~\cite{lv2025rethinking}: It compresses sequences into a small set of learned tokens per modality and performs cross-attention among tokens (plus a time-shared module) for efficient, salient fusion that tolerates time-misaligned streams.
(3) \textit{DEEPAV}~\cite{mo2024unveiling}: This method use lightweight per-modality streams followed by late cross-modal interaction and an agreement/consensus head, emphasizing decision-level fusion and robust fallback when one modality is weak or missing.

\begin{table*}[ht]
\centering
\setlength\tabcolsep{2 pt}
\footnotesize
\caption{Performance comparison across datasets (Accuracy ± Std) under asynchronous input conditions. Left block uses inverse-frequency class rebalancing (w/). Right block uses original training (w/o). Accuracy (mean$\pm$std) results are obtained on class-balanced test sets, where class balance ensures fair performance comparison without the need for F1-score reporting. The accuracy results of each single modality are calculated from each task-specific stream, while the average accuracy is simply the average of them. All models use similar capacity ($\approx$10k params; similar backbone structure). The best result is highlighted in \textbf{bold}. The second-best result is \underline{underlined}. Lower $\Delta m\%$ is better.}
\label{tab:main_results}
\begin{adjustbox}{max width=\linewidth}
\begin{tabular}{llcccccccccc}
\toprule
Dataset & Approach &
\multicolumn{5}{c}{\textbf{w/} class rebalancing} &
\multicolumn{5}{c}{\textbf{w/o} class rebalancing} \\
\cmidrule(lr){3-7}\cmidrule(lr){8-12}
& & Mod.1 & Mod.2 & Mod.3 & Avg Acc $\uparrow$ & $\Delta m\% \downarrow$
  & Mod.1 & Mod.2 & Mod.3 & Avg Acc $\uparrow$ & $\Delta m\% \downarrow$ \\
\midrule

\multirow{12}{*}{\begin{tabular}[c]{@{}c@{}}\textit{FOG}\\ \textit{Dataset}\end{tabular}}
& \multicolumn{1}{l}{\textbf{}} & \multicolumn{1}{c}{\textbf{Skeleton}} & \multicolumn{1}{c}{\textbf{IMU}} & \multicolumn{1}{c}{\textbf{—}} & \multicolumn{1}{c}{\textbf{}} & \multicolumn{1}{c}{\textbf{}} & \multicolumn{1}{c}{\textbf{Skeleton}} & \multicolumn{1}{c}{\textbf{IMU}} & \multicolumn{1}{c}{\textbf{—}} & \multicolumn{1}{c}{\textbf{}} \\ 
\cmidrule(lr){3-5}
\cmidrule(lr){8-10} 
& Skeleton Only              & \underline{$58.74\pm8.65$} & --             & -- & --                & --
                              & \underline{$47.73\pm6.07$} & --             & -- & --                & -- \\
& Sensor Only                & --             & $50.50\pm3.45$ & -- & --                & --
                              & --             & $33.47\pm0.14$ & -- & --                & -- \\
\cmidrule(r){2-12}
& Early Fusion               & $41.59\pm3.53$ & $53.41\pm8.39$ & -- & $47.50\pm2.96$    & $11.72$
                              & $41.39\pm3.72$ & $35.55\pm2.28$ & -- & $38.47\pm1.73$    & $3.53$ \\
& Late Fusion                & $44.48\pm4.50$ & $50.70\pm4.41$ & -- & $47.59\pm2.50$ & $12.21$
                              & $44.16\pm2.78$ & $36.25\pm1.72$ & -- & \underline{$40.20\pm1.74$} & \underline{$-0.41$} \\
& Shared Latent              & $37.81\pm2.40$ & $53.35\pm6.08$ & -- & $45.58\pm3.07$    & $14.99$
                              & $40.94\pm3.44$ & \underline{$36.66\pm2.43$} & -- & $38.80\pm2.10$    & $2.35$ \\
& Cross Attention            & $38.80\pm3.43$ & $48.33\pm6.08$ & -- & $43.57\pm3.11$    & $19.12$
                              & $38.99\pm1.78$ & $36.27\pm2.15$ & -- & $37.63\pm1.41$    & $4.97$ \\
& FOCAL \cite{liu2023focal}  & $48.59\pm5.09$ & \bm{$53.91\pm6.19$} & -- & $51.26\pm3.88$    & $5.26$
                              & $42.66\pm3.21$ & $34.81\pm1.74$ & -- & $38.73\pm1.68$    & $3.31$ \\
& TACA \cite{lv2025rethinking} & $45.27\pm3.02$ & $39.02\pm2.83$ & -- & $42.14\pm2.43$    & $22.83$
                              & $39.07\pm1.52$ & $33.39\pm0.16$ & -- & $36.23\pm0.81$    & $9.19$ \\
& DEEPAV \cite{mo2024unveiling} & $53.69\pm9.00$ & \underline{$53.47\pm1.26$} & -- & \underline{$54.03\pm4.46$}    & \underline{$1.36$}
                              & $45.35\pm2.84$ & $34.37\pm1.64$ & -- & $39.86\pm2.14$    & $1.15$ \\
& \cellcolor{mygrey}\texttt{TRIP} & \cellcolor{mygrey}\bm{$63.61\pm3.86$} & \cellcolor{mygrey}$49.76\pm4.97$ & \cellcolor{mygrey}-- & \cellcolor{mygrey}\bm{$56.68\pm4.17$} & \cellcolor{mygrey}\bm{$-3.41$}
                              & \cellcolor{mygrey}\bm{$63.61\pm3.86$} & \cellcolor{mygrey}\bm{$49.76\pm4.97$} & \cellcolor{mygrey}-- & \cellcolor{mygrey}\bm{$56.68\pm4.17$} & \cellcolor{mygrey}\bm{$-40.97$} \\

\midrule
\multirow{12}{*}{\begin{tabular}[c]{@{}c@{}}\textit{FOG}\\ \textit{Dataset}\end{tabular}}
& \multicolumn{1}{l}{\textbf{}} & \multicolumn{1}{c}{\textbf{Skeleton}} & \multicolumn{1}{c}{\textbf{vGRF}} & \multicolumn{1}{c}{\textbf{—}} & \multicolumn{1}{c}{\textbf{}} & \multicolumn{1}{c}{\textbf{}} & \multicolumn{1}{c}{\textbf{Skeleton}} & \multicolumn{1}{c}{\textbf{vGRF}} & \multicolumn{1}{c}{\textbf{—}} & \multicolumn{1}{c}{\textbf{}} \\ 
\cmidrule(lr){3-5}
\cmidrule(lr){8-10} 
& Skeleton Only              & $53.31\pm4.26$ & --             & -- & --                & --
                              & $41.80\pm4.46$ & --             & -- & --                & -- \\
& Sensor Only                & --             & \bm{$68.78\pm3.04$} & -- & --                & --
                              & --             & $60.13\pm3.12$ & -- & --                & -- \\
\cmidrule(r){2-12}
& Early Fusion               & $39.94\pm3.35$ & $62.50\pm3.04$ & -- & $51.22\pm1.38$    & $17.11$
                              & $35.27\pm1.41$ & $63.32\pm4.37$ & -- & $49.30\pm2.39$    & $5.16$ \\
& Late Fusion                & $46.18\pm4.03$ & $61.70\pm4.38$ & -- & $53.94\pm0.64$    & $11.83$
                              & $41.52\pm3.41$ & $63.64\pm2.81$ & -- & \underline{$52.59\pm1.71$}    & $-2.58$ \\
& Shared Latent              & $40.39\pm2.07$ & $63.18\pm2.33$ & -- & $51.79\pm0.64$    & $16.19$
                              & $35.03\pm1.66$ & \underline{$64.49\pm3.55$} & -- & $49.76\pm1.91$    & $4.47$ \\
& Cross Attention            & $34.51\pm1.21$ & $57.66\pm3.79$ & -- & $46.08\pm2.35$    & $25.72$
                              & $33.47\pm0.15$ & $54.82\pm5.49$ & -- & $44.14\pm2.75$    & $14.38$ \\
& FOCAL \cite{liu2023focal}  & $54.08\pm4.55$ & $62.07\pm3.71$ & -- & $58.07\pm3.52$    & $4.16$
                              & $51.54\pm4.59$ & $50.12\pm5.95$ & -- & $50.83\pm4.04$    & $-3.33$ \\
& TACA \cite{lv2025rethinking} & \underline{$56.32\pm1.87$} & $49.70\pm2.17$ & -- & $53.01\pm1.32$    & $11.04$
                              & \underline{$59.61\pm2.62$} & $37.62\pm0.99$ & -- & $48.62\pm1.57$    & $-2.59$ \\
& DEEPAV \cite{mo2024unveiling} & \bm{$63.58\pm6.32$} & $52.74\pm1.91$ & -- & \underline{$58.16\pm3.10$} & \bm{$2.03$}
                              & \bm{$62.12\pm3.00$} & $42.41\pm3.03$ & -- & $52.27\pm1.71$    & \underline{$-9.57$} \\
& \cellcolor{mygrey}\texttt{TRIP} & \cellcolor{mygrey}$51.24\pm5.76$ & \cellcolor{mygrey}\underline{$67.71\pm3.46$} & \cellcolor{mygrey}-- & \cellcolor{mygrey}\bm{$59.48\pm3.03$} & \cellcolor{mygrey}\underline{$2.72$}
                              & \cellcolor{mygrey}$51.24\pm5.76$ & \cellcolor{mygrey}\bm{$67.71\pm3.46$} & \cellcolor{mygrey}-- & \cellcolor{mygrey}\bm{$59.48\pm3.03$} & \cellcolor{mygrey}\bm{$-17.59$} \\
\midrule
\multirow{12}{*}{\begin{tabular}[c]{@{}c@{}}\textit{WearGait}\\ \textit{Dataset}\end{tabular}}
& \multicolumn{1}{l}{\textbf{}} & \multicolumn{1}{c}{\textbf{Walkway}} & \multicolumn{1}{c}{\textbf{Insole}} & \multicolumn{1}{c}{\textbf{IMU}} & \multicolumn{1}{c}{\textbf{}} & \multicolumn{1}{c}{\textbf{}} & \multicolumn{1}{c}{\textbf{Walkway}} & \multicolumn{1}{c}{\textbf{Insole}} & \multicolumn{1}{c}{\textbf{IMU}} & \multicolumn{1}{c}{\textbf{}} & \multicolumn{1}{c}{\textbf{}} \\
\cmidrule(lr){3-5}
\cmidrule(lr){8-10} 
& Walkway Only               & $66.22\pm1.61$ & --             & --             & --                & --
                              & $63.07\pm2.94$ & --             & --             & --                & -- \\
& Insole Only                & --             & \underline{$59.55\pm1.39$} & --             & --                & --
                              & --             & \underline{$59.23\pm1.03$} & --             & --                & -- \\
& IMU Only                   & --             & --             & \underline{$77.56\pm1.34$} & --                & --
                              & --             & --             & \underline{$77.30\pm1.41$} & --                & -- \\
\cmidrule(r){2-12}
& Early Fusion               & $54.31\pm1.28$ & $56.25\pm1.42$ & $55.19\pm0.92$ & $55.25\pm0.92$    & $17.46$
                              & $52.58\pm1.38$ & $55.97\pm0.82$ & $53.55\pm1.24$ & $53.94\pm3.66$    & $17.62$ \\
& Late Fusion                & $62.01\pm3.37$ & $54.63\pm1.60$ & $65.32\pm1.84$ & $60.65\pm1.57$ & $10.13$
                              & $54.64\pm2.14$ & $54.77\pm1.80$ & $62.69\pm2.31$ & $57.37\pm0.95$ & $13.27$ \\
& Shared Latent              & $59.98\pm1.70$ & $55.78\pm1.55$ & $59.25\pm1.77$ & $58.34\pm0.70$    & $13.12$
                              & $51.58\pm1.20$ & $56.79\pm1.51$ & $54.38\pm2.13$ & $54.25\pm1.02$    & $17.33$ \\
& Cross Attention            & $53.24\pm0.95$ & $53.97\pm1.57$ & $52.90\pm1.30$ & $53.37\pm0.91$    & $20.26$
                              & $52.42\pm0.82$ & $52.70\pm1.63$ & $52.60\pm1.21$ & $52.58\pm0.79$    & $19.95$ \\
& FOCAL \cite{liu2023focal}  & $60.60\pm1.30$ & $55.72\pm1.71$ & $61.33\pm1.21$ & $59.22\pm0.95$    & $11.95$
                              & $50.96\pm1.13$ & $57.37\pm1.33$ & $58.18\pm3.10$ & $55.50\pm1.10$    & $15.69$ \\
& TACA \cite{lv2025rethinking} & \underline{$66.62\pm0.80$} & $54.42\pm2.01$ & $61.21\pm1.55$ & \underline{$60.75\pm0.59$} & \underline{$9.70$}
                              & \underline{$65.32\pm1.25$} & $54.43\pm1.75$ & $59.76\pm1.36$ & \underline{$59.84\pm0.78$} & \underline{$9.08$} \\
& DEEPAV \cite{mo2024unveiling} & $53.08\pm2.53$ & $56.91\pm1.98$ & $52.66\pm1.52$ & $54.21\pm1.02$    & $18.79$
                              & $52.71\pm0.63$ & $57.74\pm0.90$ & $52.90\pm1.36$ & $54.45\pm0.32$    & $16.84$ \\
& \cellcolor{mygrey}\texttt{TRIP} & \cellcolor{mygrey}\bm{$71.08\pm1.86$} & \cellcolor{mygrey}\bm{$63.03\pm1.04$} & \cellcolor{mygrey}\bm{$80.07\pm1.77$} & \cellcolor{mygrey}\bm{$71.39\pm1.09$} & \cellcolor{mygrey}\bm{$-5.47$}
                              & \cellcolor{mygrey}\bm{$71.08\pm1.86$} & \cellcolor{mygrey}\bm{$63.03\pm1.04$} & \cellcolor{mygrey}\bm{$80.07\pm1.77$} & \cellcolor{mygrey}\bm{$71.39\pm1.09$} & \cellcolor{mygrey}\bm{$-8.69$} \\
\bottomrule
\end{tabular}
\end{adjustbox}
\end{table*}

\begin{table}[ht]
  \centering
  \setlength\tabcolsep{2 pt}
  \footnotesize
  \caption{Performance of FOG and WearGait datasets (Accuracy $\pm$ Std) under synchronous input condition. Left block uses inverse-frequency class rebalance (w/). Right block uses original training (w/o). The best result is highlighted in \textbf{bold}. The second-best result is \underline{underlined}. Lower $\Delta m\%$ is better.}
  \label{tab:sync_results}
  \begin{adjustbox}{max width=\linewidth}
  \begin{tabular}{llcccc}
    \toprule
    Dataset & Approach & \multicolumn{2}{c}{\textbf{w/} class rebalancing} & \multicolumn{2}{c}{\textbf{w/o} class rebalancing} \\
    \cmidrule(lr){3-4}\cmidrule(lr){5-6}
            &         & Acc (\%) $\uparrow$ & $\Delta m\%$ $\downarrow$ & Acc (\%) $\uparrow$ & $\Delta m\%$ $\downarrow$ \\
    \midrule
    \multirow{10}{*}{\textit{FOG}}
      & Skeleton Only              & $56.95 \pm 5.10$ & –      & $56.44 \pm 7.27$ & – \\
      & IMU Only                   & $61.09 \pm 4.20$ & –      & \underline{$57.49 \pm 4.52$} & – \\
      \cmidrule(r){2-6}
      & Early Fusion               & $61.31 \pm 6.15$ & $-4.01$ & $48.54 \pm 5.55$ & $14.78$ \\
      & Late Fusion                & \underline{$62.32 \pm 5.73$} & \underline{$-5.72$} & $50.11 \pm 4.54$ & $12.03$ \\
      & Shared Latent             & $61.03 \pm 7.20$ & $-3.53$ & $45.76 \pm 3.90$ & $19.66$ \\
      & Cross Attention           & $55.86 \pm 7.84$ & $5.24$  & $47.37 \pm 4.13$ & $16.84$ \\
      & FOCAL \cite{liu2023focal} & $56.47 \pm 10.41$& $4.20$  & $46.87 \pm 4.46$ & $17.71$ \\
      & TACA \cite{lv2025rethinking} & $60.05 \pm 5.47$ & $-1.87$ & $46.27 \pm 3.71$ & $18.77$ \\
      & DEEPAV \cite{mo2024unveiling} & $61.35 \pm 7.99$ & $-4.08$ & $51.55 \pm 4.09$ & \underline{$9.50$} \\
      & \cellcolor{mygrey}\texttt{TRIP} & \cellcolor{mygrey}\bm{$62.35 \pm 4.38$} & \cellcolor{mygrey}\bm{$-5.77$} & \cellcolor{mygrey}\bm{$62.35 \pm 4.38$} & \cellcolor{mygrey}\bm{$-9.46$} \\
    \midrule
    \multirow{12}{*}{\textit{WearGait}}
      & Walkway Only              & $66.59 \pm 1.63$ & –       & $64.01 \pm 2.10$ & – \\
      & Insole Only               & $59.77 \pm 0.91$ & –       & $59.59 \pm 1.11$ & – \\
      & IMU Only                  & $77.14 \pm 2.23$ & –       & $76.60 \pm 2.49$ & – \\
      \cmidrule(r){2-6}
      & Early Fusion              & \underline{$82.55 \pm 1.64$} & \underline{$-23.03$} & \underline{$81.88 \pm 6.06$} & \underline{$-24.07$} \\
      & Late Fusion               & $72.28 \pm 1.69$ & $-7.72$ & $69.44 \pm 2.65$ & $-5.22$ \\
      & Shared Latent            & $60.86 \pm 2.20$ & $9.30$  & $56.53 \pm 0.64$ & $14.34$ \\
      & Cross Attention          & $77.36 \pm 0.97$ & $-15.30$& $79.30 \pm 1.58$ & $-20.16$ \\
      & FOCAL \cite{liu2023focal}& $72.92 \pm 0.83$ & $-8.68$ & $71.78 \pm 1.58$ & $-8.77$ \\
      & TACA \cite{lv2025rethinking} & $58.39 \pm 0.78$ & $12.98$ & $74.04 \pm 4.69$ & $-12.19$ \\
      & DEEPAV \cite{mo2024unveiling} & $76.46 \pm 1.61$ & $-13.95$& $50.74 \pm 3.14$ & $23.11$ \\
      & \cellcolor{mygrey}\texttt{TRIP} & \cellcolor{mygrey}\bm{$84.18 \pm 1.84$} & \cellcolor{mygrey}\bm{$-25.46$} & \cellcolor{mygrey}\bm{$84.18 \pm 1.84$} & \cellcolor{mygrey}\bm{$-28.17$} \\
    \bottomrule
  \end{tabular}
  \end{adjustbox}
\end{table}

\section{Evaluation}
\subsection{Evaluation Metric}
In addition to reporting individual performance and mean accuracy, we also incorporate a widely used metric, $\Delta m\%$~\cite{maninis2019attentive}, which evaluates the overall degradation compared to independently trained models that are considered as the reference oracles. The formal definition of $\Delta m\%$ is given as: 
\begin{align}
 \Delta m\% = \frac{1}{M}\sum_{i=1}^M (-1)^{\delta _i}(P_{m,i} - P_{b,i}) / P_{b,i}   
\end{align}
where $P_{m,i}$ and $P_{b,i}$ represent the metric $P_i$ for the compared method and the single modality-based model, respectively. The value of $\delta _i$ is assigned as 1 if a higher value is better for $P_i$, and 0 otherwise. 

\subsection{Parkinson's Disease Multimodal Datasets}
\label{sec:datasets}

We conducted evaluations on three publicly available multimodal PD gait datasets.
All datasets contain gait data from two to three complementary modalities and accompanied by clinically annotated ratings. These characteristics make them suitable for the evaluation of our proposed framework.

\smallskip
\textbf{FOG Dataset} \cite{ribeiro2022public}: It contains video and IMU recordings from 35 PD subjects performing turning-in-place at 30 Hz (video) and 128 Hz (IMU) with clinical FoG annotations. Coordinates of 3D poses are obtained using MMPose \cite{mmpose2020} (MotionBERT trained on Human3.6M).
\textbf{FBG Dataset} \cite{shida2023public}: It comprises synchronized full-body kinematics and kinetics from 26 PD subjects walking overground in ON/OFF medication states, including 3D motion capture at 150 Hz and force-plate data, with MDS-UPDRS III and H\&Y labels. 
\textbf{WearGait Dataset} \cite{kontson2024wearables}: It provides synchronized IMU, sensorized insole (16-sensor plantar pressures and embedded inertial signals), and walkway data, all at 100Hz, from 98 people with PD and 83 age-matched controls with clinical metadata. 

\subsection{Overall Evaluation}

We evaluate the proposed \texttt{TRIP} framework under both asynchronous and synchronous conditions, each with and without class re-balancing, to comprehensively assess robustness and generalization. Single-modality oracles (e.g., Skeleton Only) serve as upper bounds, while $\Delta m\%$ quantifies how multimodal fusion compares to its strongest unimodal counterpart (negative indicates improvement).

\smallskip
\noindent\textbf{Asynchronous Inputs.}
Under naturally imbalanced, asynchronous inputs—typical in clinical deployment—\texttt{TRIP} consistently achieves the highest accuracies (right blocks in Table~\ref{tab:main_results}).
For FOG, \texttt{TRIP} attains 56.68\%, surpassing the best baseline (Late Fusion, 40.2\%) and yielding a large multimodal gain ($\Delta m$=$-$41.97). Skeleton accuracy rises from 47.73\% to 63.61\% (+15.88 pp), and IMU from 33.47\% to 49.76\% (+16.29 pp).
For FBG, it reaches 59.48\%, outperforming DEEPAV (52.59\%) by +6.89 pp, with Skeleton improved from 41.8\% to 51.24\% (+9.44 pp) and vGRF from 60.13\% to 67.71\% (+7.58 pp), resulting in $\Delta m$=$-$17.59.
For WearGait, which includes three modalities, \texttt{TRIP} achieves 71.39\%, outperforming all fusion methods while maintaining strong per-modality accuracies (Walkway 71.08\%, Insole 63.03\%, IMU 80.07\%). These results confirm that \texttt{TRIP} can effectively learn from time-misaligned data without synchronized sampling, while other methods deteriorate.

\smallskip
\noindent\textbf{Synchronous Inputs.}
When all modalities are synchronized (Table~\ref{tab:sync_results}), \texttt{TRIP} retains its lead (right blocks).
On FOG, it achieves 62.35\%, outperforming all fusion baselines and exceeding the best unimodal oracle (57.49\%) by +4.86 pp ($\Delta m$=$-$9.46).
On WearGait, \texttt{TRIP} reaches 84.18\%, the best among all models, with a substantial multimodal gain ($\Delta m$=$-$28.17). It surpasses all fusion competitors (e.g., Early Fusion 81.88\%, Cross Attention 79.3\%) and all single-modality oracles (Walkway 64.01\%, Insole 59.59\%, IMU 76.6\%).
We omit FBG synchronous results since inconsistent temporal alignment across sensor trials prevents fair synchronized evaluation, further underscoring the limitations of sync-dependent approaches.

\smallskip
\noindent\textbf{Effect of Class Rebalancing.}
To confirm that improvements are not caused by label imbalance, we repeat experiments with inverse-frequency weighting on the loss function for all baseline methods.
Under \textit{asynchronous + rebalanced} training (left blocks in Table~\ref{tab:main_results}), \texttt{TRIP} maintains its superiority in mean accuracy: 56.68\% on FOG ($\Delta m$=$-$3.41), 59.48\% on FBG ($\Delta m$=2.72), and 71.39\% on WearGait ($\Delta m$=$-$5.47).
Under \textit{synchronous + rebalanced} training (left blocks in Table~\ref{tab:sync_results}), \texttt{TRIP} again performs best: 62.35\% on FOG ($\Delta m$=$-$5.77) and 84.18\% on WearGait ($\Delta m$=$-$25.46).
Even when class frequencies are equalized, \texttt{TRIP} outperforms or matches the best rebalanced baselines (e.g., DEEPAV*, Early Fusion*), while preserving high per-modality accuracies.

\smallskip
\noindent\textbf{Summary.}
Across all configurations—synchronous or asynchronous and balanced or unbalanced—\texttt{TRIP} consistently delivers the strongest fusion accuracy and the lowest $\Delta m\%$. These findings confirm that:
(1) \texttt{TRIP} robustly integrates complementary modalities without temporal alignment;
(2) it enhances weaker modalities while maintaining strong ones via gradient-balanced optimization; and
(3) it remains stable under class rebalancing, demonstrating intrinsic robustness to both modality and label imbalance.

\subsection{Modality Collapse Mitigation}
\begin{figure}[t]
    \centering
    \includegraphics[height=6cm,width= \linewidth]{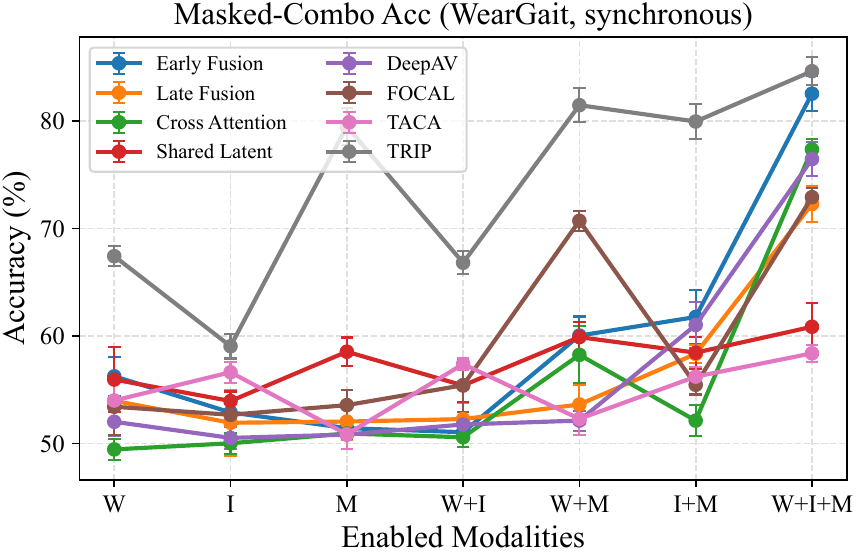}
    \caption{Demonstration of effective modality collapse mitigation with \texttt{TRIP}. Results are obtained using WearGait dataset under synchronous inputs condition. Values in x-axis include all combinations of different modalities used during inference (W: Walkway, I: Insole, M: IMU).}
    \label{fig:mod_collapse}
\end{figure}

Apart from effective multimodal information fusion, compared to all other fusion baselines, \texttt{TRIP} keeps much higher single and pairwise (i.e., combine two modalities) accuracies when we zero-mask other streams at test time. As shown in Fig. \ref{fig:mod_collapse}, most of other baselines' results hover around 50–60\% under the same masks. In addition, pairwise drops from full are smaller, indicating graceful degradation rather than failure when one modality is missing. That means \texttt{TRIP} learned useful per-stream representations instead of collapsing to one single modality.


\begin{table}[ht]
  \centering
  \footnotesize
  \caption{Ablation study on important modules/loss functions on three datasets (Accuracy $\pm$ Std) with asynchronous inputs. The best result is highlighted in \textbf{bold}. The second-best result is \underline{underlined}.}
  \label{tab:ablation_study}
  \begin{adjustbox}{max width=\linewidth}
    \begin{tabular}{cccc}
      \hline
      \toprule
      Dataset & MOO & Margin Rebalancing & Accuracy (\%) $\uparrow$ \\
      \midrule
      \multirow{4}{*}{\begin{tabular}[c]{@{}c@{}}\textit{FOG}\\\textit{Dataset}\end{tabular}}
        &             &               & $ 35.25 \pm 1.26 $  \\
        & \ding{51}  &                & $ 36.32 \pm 2.39 $  \\
        &             & \ding{51}     & \underline{$ 52.01 \pm 2.01 $}  \\ 
        & \ding{51}  & \ding{51}      & \bm{$ 56.68 \pm 4.17 $}  \\
        \midrule
      \multirow{4}{*}{\begin{tabular}[c]{@{}c@{}}\textit{FBG}\\\textit{Dataset}\end{tabular}}
        &             &                & $36.20 \pm 0.75 $  \\
        & \ding{51}  &                 & $48.78 \pm 1.85 $  \\
        &             & \ding{51}      & \underline{$53.66 \pm 1.85 $}  \\ 
        & \ding{51}  &  \ding{51}      & \bm{$59.48 \pm 3.03$}   \\
        \midrule
      \multirow{4}{*}{\begin{tabular}[c]{@{}c@{}}\textit{WearGait}\\\textit{Dataset}\end{tabular}}
        &           &            & $ 57.38 \pm 0.95 $  \\
        & \ding{51} &            & $ 69.95 \pm 0.74 $  \\
        &           & \ding{51}  & \underline{$ 70.86 \pm 1.17 $}  \\ 
        & \ding{51} & \ding{51}  & \bm{$ 71.39 \pm 1.09 $}   \\
      \bottomrule
      \hline
    \end{tabular}
  \end{adjustbox}
\end{table}
\subsection{Ablation Study}
Our approach has two key pieces: a multi-objective optimization (MOO) paradigm and a margin-based class rebalancing strategy. Table~\ref{tab:ablation_study} reports an ablation with asynchronous inputs across three datasets. Both components consistently improve accuracy, and the best results arise when they are combined (bold). MOO is especially impactful when class imbalance is milder (FBG and WearGait), delivering gains of 12.58\% and 12.57\%, respectively, over the non-MOO counterpart. While each component alone outperforms the plain baseline, only their combination achieves the top performance across datasets (with the second-best underlined), highlighting the complementary nature of MOO and margin-based rebalancing.

\begin{figure}[t]
    \centering
    \includegraphics[height=4.5cm,width=\linewidth]{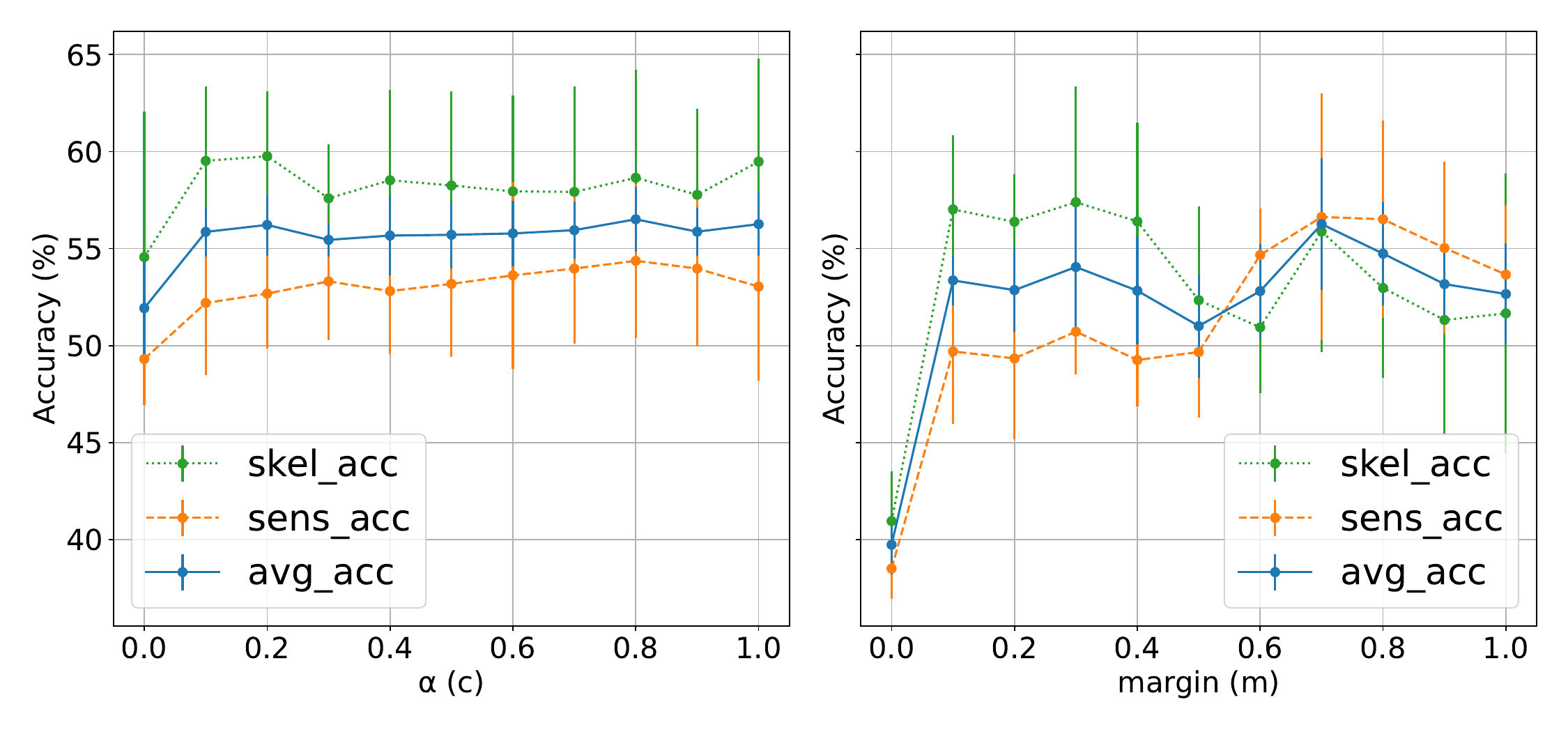}
    \caption{Performances analysis under asynchronous input settings on the FOG dataset with two hyperparameters: $\alpha$ and $m$, which respectively represents the MOO coefficient and additive margin coefficient.}
    \label{fig:hyperparameters_analysis}
\end{figure}

\subsection{Hyper-Parameters Analysis}
Figure \ref{fig:hyperparameters_analysis} reports the mean accuracy obtained when varying the two free hyper-parameters of our framework:  
1. the MOO coefficient $\alpha$, which dictates how strongly the shared backbone is steered toward the worst-performing modality, and  
2. the additive margin $m$ used in our class-rebalancing loss.

\paragraph{MOO Coefficient $\alpha$} The left plot shows a monotonic but saturating trend: raising $\alpha$ from $0$ (plain averaging) to $\approx0.8$ yields a gradual $\sim\!2$–3 pp gain in average accuracy, mainly by lifting the weaker sensor branch while leaving the skeleton branch essentially unchanged. Beyond $\alpha\!\approx\!0.9$ the curve flattens and variance grows, suggesting that very aggressive worst-case weighting offers no further benefit. Hence a moderate range $\alpha\in[0.6,0.9]$ is sufficient for stable improvements.

\paragraph{Additive Margin Coefficient $m$} The right plot highlights a sharp jump in performance when a small margin is introduced: moving from $m=0$ to $m\!\approx\!0.3$ boosts all branches, confirming that class-adaptive margins effectively compensate for long-tailed label distributions. Larger margins ($m>0.5$) lead to oscillations—sensor accuracy continues to rise slightly, whereas skeleton accuracy deteriorates—indicating that excessive separation can impede convergence of the harder (minority) classes.

\paragraph{Recommended Setting} Taken together, the analysis shows that the two mechanisms act \emph{orthogonally}: $\alpha$ resolves inter-modality gradient conflict, while $m$ improves intra-class discrimination. Empirically, $\alpha\!\approx\!0.7$–$0.9$ and $m\!\approx\!0.3$–$0.5$ strike a good balance between efficacy and stability across datasets.


\section{Conclusion and Discussion}


In this paper, we approach PD assessment from the perspective of MOO, aiming to facilitate more practical deployment of AI-assisted solutions in real-world scenarios. Extensive experiments demonstrate that our proposed framework, \texttt{TRIP}, not only supports flexible input configurations during both training and inference but also achieves competitive performance. Nevertheless, \texttt{TRIP} has certain limitations—for instance, it introduces additional hyper-parameters that may require careful tuning. We plan to further address these challenges in our future work.

\newpage

 





\end{document}